\documentclass{article}

\usepackage{graphicx}

\usepackage{amsmath}
\usepackage{amsfonts}
\usepackage{amssymb}
\usepackage{algorithm}
\usepackage{algorithmic}
\usepackage{hyperref}

\newtheorem{definition}{Definition}[section]
\newtheorem{proposition}{Proposition}[section]

\DeclareMathOperator*{\argmax}{arg\,max}

\begin{document}

\title{Learning Non-Markovian Reward Models\\ in MDPs}

\author{
Gavin Rens\\
KU Leuven, Belgium, email: gavin.rens@cs.kuleuven.be\\[7pt]
Jean-Fran\c{c}ois Raskin\\
Universit\'e Libre de Bruxelles, Belgium, email: jraskin@ulb.ac.be
}

\maketitle

\begin{abstract}
There are situations in which an agent should receive rewards only after having accomplished a series of previous tasks. In other words, 
the reward that the agent receives is {\em non-Markovian}.
One natural and quite general way to represent history-dependent rewards is via a {\em Mealy machine}; 
a finite state automaton that produces output sequences (rewards in our case) from input sequences (state/action observations in our case).
In our formal setting, we consider a Markov decision process (MDP) that models the dynamic of the environment in which the agent evolves and a Mealy machine synchronised with this MDP to formalise the non-Markovian reward function. While the MDP is known by the agent, the reward function is unknown from the agent and must be learnt.

Learning non-Markov reward functions is a challenge. Our approach to overcome this challenging problem is a careful combination of the Angluin's $L^*$ active learning algorithm to learn finite automata, testing techniques for establishing conformance of finite model hypothesis and optimisation techniques for computing optimal strategies in Markovian (immediate) reward MDPs.
We also show how our framework can be combined with classical heuristics such as Monte Carlo Tree Search.
We illustrate our algorithms and a preliminary implementation on two typical examples for AI.
\end{abstract}

\section{Introduction}

Traditionally, a Markov Decision Process (MDP) models the probability distribution to go to a state $s'$ from  the current state $s$ while taking a given action $a$ together with an {\em immediate reward} that is received while performing $a$ from $s$.
This immediate reward is defined regardless of the history of states traversed in the past. This immediate reward has thus the {\em Markovian property}. But many situations  require the reward to depend on the history of states visited so far. A reward may depend on the particular sequence of (sub)tasks that has been completed. For instance, when a nuclear power plant is shut down in an emergency, there is a specific sequence of operations to follow to avoid a disaster; or in legal matters, there are procedures to follow which require documents to be submitted in a particular order, etc.

Learning and maintaining non-Markovian reward functions is useful for several reasons:
$(i)$ Many tasks are described intuitively as a sequence of sub-tasks or mile-stones, each with their own reward (cf.\ Sec.~\ref{sec:Related-Work}.)
$(ii)$ Possibly, not all relevant features are available in state descriptions, or states are partially observable, making it necessary to visit several states before (more) accurate rewards can be dispensed~\cite{abz10,twkvcm19}.
$(iii)$ Automata (reward machines) are useful for modeling  desirable and undesirable situations so that tracking and predicting beneficial and detrimental situations~\cite{abeknt18,kpr18,dfip19}.
Actually, in practice, it can be argued that non-Markovian tasks are more the norm than Markovian ones.

In this work, we describe an {\em active} learning algorithm to automatically learn through experiments reward models defined as Mealy machines. A Mealy machine is a deterministic finite automaton (DFA) that produces output sequences that are rewards in our case, from input sequences that are state/action observations in our setting. We refer to such finite state reward models as  \textit{Mealy Reward Machines} (MRM).
Our algorithm is based on a careful combination of Angluin's $L^*$ active learning algorithm \cite{a87} to learn finite automata, testing techniques for establishing the conformance of (hypothesized) finite models and optimization techniques for computing optimal strategies in classical immediate reward MDPs. By using formal methods for automata inference for learning a reward machine, one can draw on the vast literature on formal methods and model-checking to obtain guarantees under precisely stated conditions.
Our contribution is to show how a non-Markovian reward function for a MDP can be actively learnt and exploited to play optimally according to this non-Markovian reward function in the MDP. We provide a framework for completely and correctly learning the underlying reward function with guarantees under mild assumptions. To the best of our knowledge, this is the first work which employs traditional automata inference for learning a non-Markovian reward model in an MDP setting.

Next, we discuss related work and then cover the necessary formal concepts and notations.
In Section~\ref{sec:Modeling-Non-Markovian-Rewards}, we define our Mealy Reward Machine (MRM), which represents an agent's non-Markovian reward model.
Section~\ref{sec:Learning-Mealy-Reward-Machines} explains how an agent can infer/learn an underlying MRM and present one method for exploiting a learnt MRM.
Section~\ref{sec:Experimental-Evaluation} reports on experiments involving learning and exploiting MRMs; we consider two scenarios.
The last section concludes this paper and points to future research directions.

\section{Related Work}
\label{sec:Related-Work}

There has been a surge of interest in non-Markovian reward functions recently, with most papers on the topic being publications since 2017. But unlike our paper, most of those papers are not concerned with {\em learning} non-Markovian reward functions.

A closely related and possibly overlapping research area is the use of temporal logic (especially linear temporal logic (LTL)), for specifying tasks in Reinforcement Learning (RL) \cite{abeknt18,tkvm18a,tkvm18b,ctkvm19,dfip19}.
Building on recent progress in temporal logics over finite traces (LTL$_f$), the authors of~\cite{bdp18} adopt linear dynamic logic on finite traces (LDL$_f$; an extension of LTL$_f$) for specifying non-Markovian rewards, and provide an automaton construction algorithm for building a Markovian model. The approach is claimed to offer strong minimality and compositionality guarantees.

An earlier publication that deserves to be mentioned is~\cite{bbg96}. In this paper, the authors propose to encode non-Markovian rewards by assigning values using temporal logic formulae. The reward for being in a state is then the value associated with a formula that is true in that state. As it is the case in the present work, they considered systems that can be modeled as MDPs, but the non-Markovian reward functions is given and does not need to be learnt. 
The authors of that paper were the first to abbreviate the class of MDP with non-Markovian reward as  \textit{decision processes with non-Markovian reward} or NMRDP for short.
In~\cite{tgspk06}, the author present the Non-Markovian Reward Decision Process Planner: a software platform for the development and experimentation of methods for decision-theoretic planning with non-Markovian rewards. 




In another paper, the authors are concerned with both the specification and effective exploitation of non-Markovian reward in the context of MDPs \cite{ccsm18}. They specify non-Markovian reward-worthy behavior in LTL. These behaviors are then translated to corresponding deterministic finite state automata whose accepting conditions signify satisfaction of the reward-worthy behavior. These automata accepting conditions form the basis of Markovian rewards that can be solved by off-the-shelf MDP planners, while preserving policy optimality guarantees. In that sense, it is similar to our framework.

None of the research mentioned above is concerned with \textit{learning} non-Markovian reward functions.
However, in the work of \cite{twkvcm19}, an agent incrementally learns a reward machine (RM) in a partially observable MDP (POMDP). 
They use a set of traces as data to solve an optimization problem. ``If at some point [the RM] is found to make incorrect predictions, additional traces are added to the training set and a new RM is learned.''
Each trace is of the form
$
o_1 a_1 r_1 \cdots o_k a_{k} r_{k},
$
where the $o_i$ are observations (as required for POMDP models).
An RM is specified as the constraints of the optimization model. They use \textit{Tabu search} to solve the optimization problem. 
They have a theorem stating ``When the set of training traces (and their lengths) tends to infinity [...], any perfect RM with respect to $L$ and at most $u_\mathit{max}$ states will be an optimal solution to the formulation LRM.'' Here $L$ is a labeling function which assigns truth values to propositional symbols given an observation, action and next observation, and $u_\mathit{max}$ is an upper bound on the number of states in an RM. A perfect RM is simply an RM that makes the correct predictions for a given POMDP.
Their approach is also active learning: If on any step, there is evidence that the current RM might not be the best one, their approach will attempt to find a new one. One strength of their method is that the the reward machine is constructed over a set of propositions, where propositions can be combined to represent transition/rewarding conditions in the machine. Currently, our approach can take only single observations as transition/rewarding conditions.
However, they do not consider the possibility to compute optimal strategies using \textit{model-checking techniques}. 

Moreover, our approach is different to that of Toro Icarte et al. \cite{twkvcm19} in that ours is an active learning approach, where the agent is guided by the L* algorithm to answer exactly the queries required to find the underlying reward machine. It seems that the approach of Toro Icarte et al. does not have this guidance and interaction with the learning algorithm.

\section{Formal Preliminaries}

We review Markov Decision Processes (MDPs)
and Angluin-style learning of Mealy machines.


An (immediate-reward) MDP is a tuple $\langle S, A, T, R, s_0 \rangle$, where
\begin{itemize}
\item  $S$ is a finite set of states,
\item $A$ is a finite set of actions,
\item $T:S\times A\times S\mapsto [0,1]$ is the state transition function such that $T(s,a,s')$ is the probability that action $a$ causes a system transition from state $s$ to state $s'$,
\item $R:A\times S\mapsto \mathbb{R}$ is the reward function such that $R(a,s)$ is the immediate rewards for performing action $a$ in state $s$, and
\item $s_0$ the initial state the system is in.
\end{itemize}
A non-rewarding MDP (nrMDP) is a tuple $\langle S, A, T, s_0 \rangle$ without a reward function.

The following description is from \cite{v17}.
Angluin \cite{a87} showed that finite automata can be learned using the so-called membership and equivalence queries.
Her approach views the learning process as a game between a \textit{minimally adequate teacher} (MAT) and a learner who wants to discover the automaton. In this framework, the learner has to infer the input/output behavior of an unknown Mealy machine by asking queries to a teacher. The MAT knows the Mealy machine $\mathcal{M}$.
Initially, the learner only knows the inputs $\mathcal{I}$ and outputs $\mathcal{O}$ of $\mathcal{M}$. The task of the learner is to learn $\mathcal{M}$ through two types of queries:
\begin{itemize}
\item
With a \textit{membership query} (MQ), the learner asks what the output is in response to an input sequence
$\sigma \in \mathcal{I}^*$. The teacher answers with output sequence $\mathcal{M}(\sigma)$.
\item
With an \textit{equivalence query} (EQ), the learner asks if a hypothesized Mealy machine $\mathcal{H}$ with inputs $\mathcal{I}$ and outputs $\mathcal{O}$ is correct, that is, whether $\mathcal{H}$ and $\mathcal{M}$ are equivalent ($\forall \sigma\in \mathcal{I}^*, \mathcal{M}(\sigma) = \mathcal{H}(\sigma)$). The teacher answers \textit{yes} if this is the case. Otherwise she answers \textit{no} and supplies a \textit{counter-example} $\sigma' \in \mathcal{I}^*$ that distinguishes $\mathcal{H}$ and $\mathcal{M}$ (i.e., such that $\mathcal{M}(\sigma') \neq \mathcal{H}(\sigma)$).
\end{itemize}

The $L^*$ algorithm incrementally constructs an \textit{observation table} with entries being elements from $\mathcal{O}$.

Two crucial properties of the observation table allow for the construction of a Mealy machine \cite{v17}: closedness and consistency.
We call a closed and consistent observation table \textit{complete}.

Angluin \cite{a87} proved that her $L^*$ algorithm is able to learn a finite state machine (incl. a Mealy machine) by asking a polynomial number of membership and equivalence queries (polynomial in the size of the corresponding minimal Mealy machine equivalent to $\mathcal{M}$).
Let $|\mathcal{I}|$ be the size of the input alphabet (observations), $n$ be the total number of states in the target Mealy machine, and $m$ be the maximum length of the counter-example provided for learning the machine. Then the correct machine can be produced by asking maximum $O(|\mathcal{I}|^2 + |\mathcal{I}|mn^2)$ queries (using, e.g., the ${L_M}^+$ algorithm) \cite{sg09}.

In an ideal (theoretical) setting, the agent (learner) would ask a teacher whether $\mathcal{H}$ is correct (equivalence query), but in practice, this is typically not possible \cite{v17}.
``Equivalence query can be approximated using a conformance testing (CT) tool \cite{ly96} via a finite number of test queries (TQs). A test query asks for the response of the [system under learning] to an input sequence, similar to a membership query. If one of the test queries exhibits a counterexample then the answer to the equivalence query is \textit{no}, otherwise the answer is \textit{yes}'' \cite{v17}. Vaandrager \cite{v17} cites Lee and Yannakakis \cite{ly96} saying that a finite and complete conformance test suite does exist if we assume a bound on the number of states of a Mealy machine.

\section{Modeling Non-Markovian Rewards}
\label{sec:Modeling-Non-Markovian-Rewards}

\begin{figure}[t]
\centering
\includegraphics[scale=0.3]{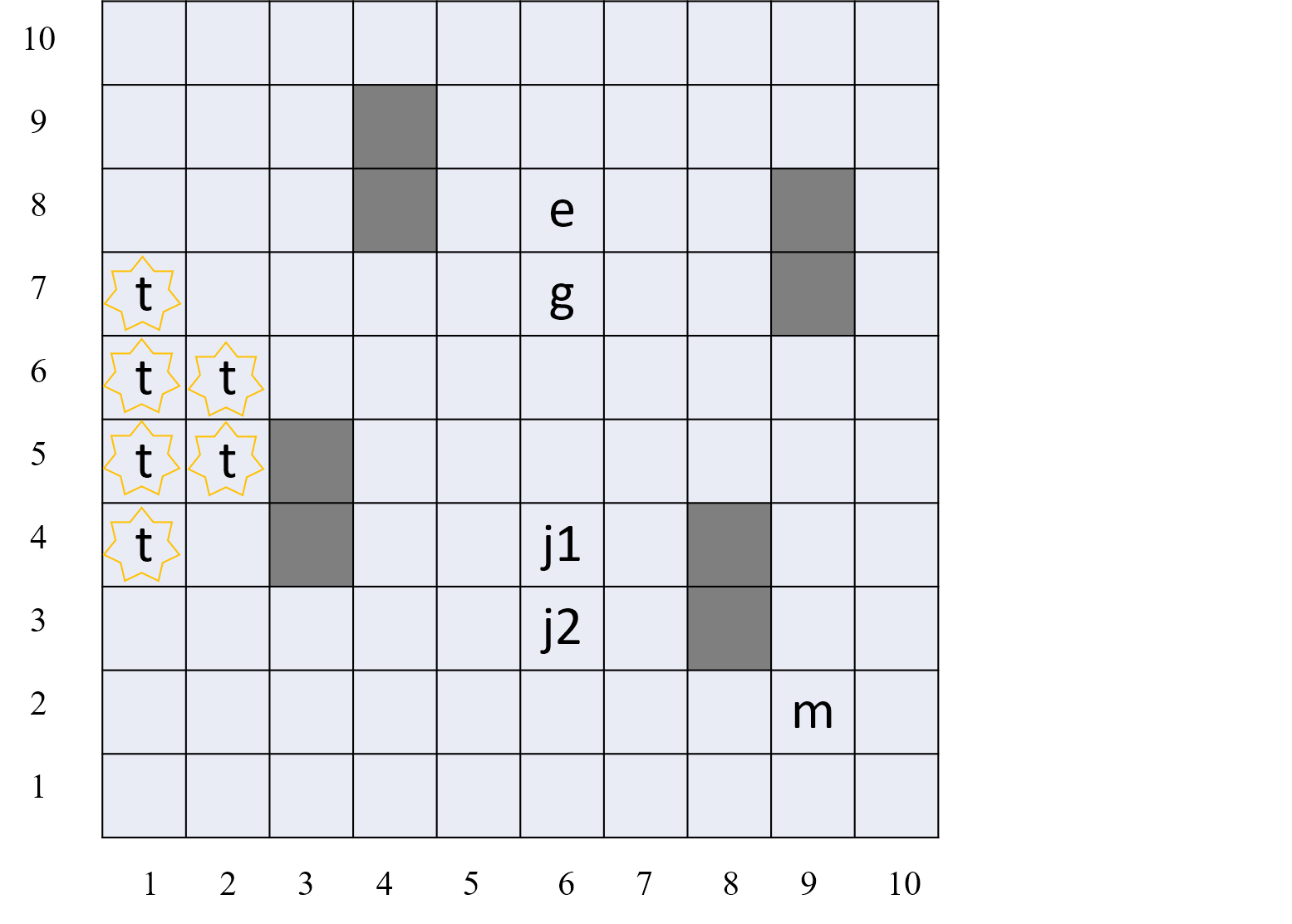}
\caption{\label{fig:scenario}
The agent has a task of finding the person with the treasure-map at \textsf{m}, then either (i) selling the map to a jeweller at \textsf{j1} or \textsf{j2} or (ii) collecting equipment at \textsf{e} or hiring a guide at \textsf{g}, then finding the treasure at \textsf{t}, and finally selling the treasure to one of the jewelers. The agent can continue process (ii) once it has the map. Default reward/cost $c=0$. Blank cells contain observation $null$ by default.} 
\end{figure}

\begin{figure}[t]
\centering
\includegraphics[scale=0.3]{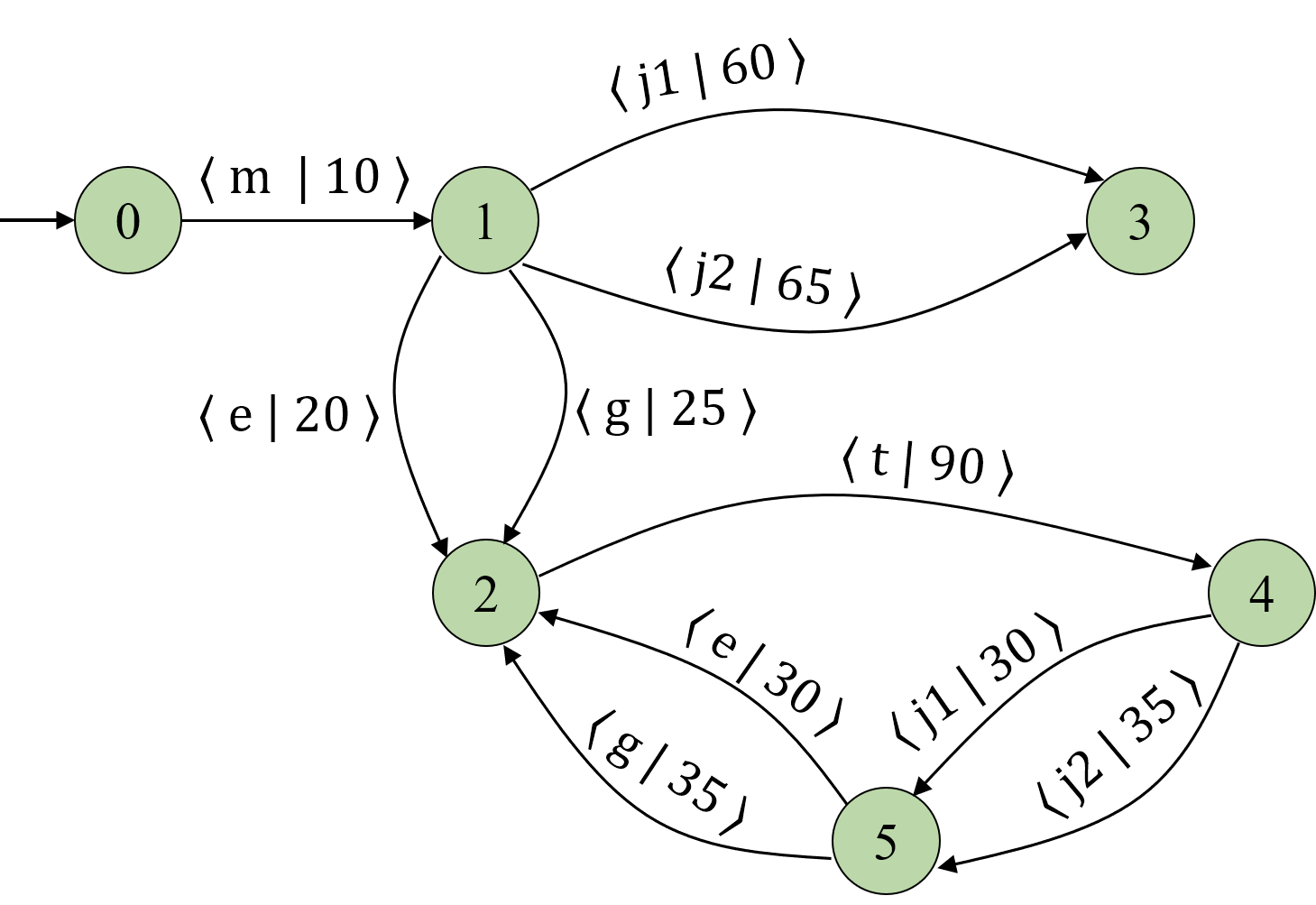}
\caption{\label{fig:MRM}
A Mealy reward machine for the treasure map scenario. In the case that $null$ is in the input alphabet, every node has a self-loop with label $\langle null \mid c\rangle$; not shown.}
\end{figure}

\paragraph{Running Example.}
Consider, as a running example, an agent who might stumble upon a person with a map for a hidden treasure and some instruction on how to retrieve the treasure. The instructions imply that the agent purchase some specialized equipment before going to the cave marked on the treasure map. Alternatively, the agent may hire a guide who already has the necessary equipment. If the agent is lucky enough to find some treasure, the agent must sell it at the local jewelry traders. There are two jewelers. The agent can then restock its equipment or re-hire a guide, get some more treasure, sell it and so on. The agent could also sell the map to one of the jewelers without looking for the treasure. Unfortunately, the instructions are written in a coded language which the agent cannot read. However, the agent recognizes that the map is about a hidden treasure, and thus spurs the agent on to start treasure hunting to experience rewards and learn the missing information.

The reward behavior in this scenario is naturally modeled as an automaton where the agent receives a particular reward given a \textit{particular sequence} of observations. There is presumably higher value in purchasing equipment for treasure hunting only \textit{after} coming across a treasure map and thus deciding to look for a treasure. There is more value in being at the treasure with the right equipment than with no equipment, etc.

We shall interpret features of interest as observations: (obtaining) a map, (purchasing) equipment, (being at) a treasure, (being at) a jewelry trader, for example.
Hence, for a particular sequence of input observations, the Mealy machine outputs a corresponding sequence of rewards. If the agent tracks its observation history, and if the agent has a correct automaton model, then it knows which reward to expect (the last symbol in the output) for its observation history as input. 
Figure~\ref{fig:scenario} depicts the scenario in two dimensions.
The underlying Mealy machine could be represented graphically as in Figure~\ref{fig:MRM}. It takes observations as inputs and supplies the relevant outputs as rewards.
For instance, if the agent sees $\mathtt{m}$, (map), then $\mathtt{j1}$, then $\mathtt{j2}$, then the agent will receive rewards 10, then 60 and then 0. And if it sees the sequence $\mathtt{m} \cdot\mathtt{g} \cdot\mathtt{t} \cdot \mathtt{j2}$, then it will receive the reward sequence $10 \cdot 25\cdot 90\cdot 35$.

We define \textit{intermediate states} as states that do not signify a significant progress towards completion of a task. In Figure~\ref{fig:scenario}, all lighter-colored blank cells represent intermediate states.
We assume that there is a default reward/cost an agent gets for entering intermediate states. This default reward/cost is fixed and denoted $c$ in general discussions. Similarly, the special null observation ($null$) is observed in intermediate states. An agent might or might not be designed to recognize intermediate states. 
If the agent cannot distinguish intermediate states from `significant' states, then $null$ will be treated exactly like all other observations and it will have to learn transitions for $null$ observations (or worse, for all observations associated with intermediate states). If the agent \textit{can} distinguish intermediate states from `significant' states, then we can design methods to take advantage of this background knowledge. Our approximate active learning algorithm (Alg.~\ref{alg:approximate-active-learning}) is an example of how the ability to recognize intermediate states can be taken advantage of.

\paragraph{Mealy Reward Machines}

We introduce the Mealy Reward Machine to model non-Markovian rewards. These machines take a set $Z$ of observations representing high-level features that the agent can detect (equivalent to the set of input symbols for $L^*$). A labeling function $\lambda:A\times S \mapsto Z\uplus \{null\}$ maps action-state pairs onto observations; $S$ is the set of nrMDP states and $Z$ is a set of observations. The meaning of $\lambda(a,s)= z$ is that $z$ is observed in state $s$ reached via action $a$. For Mealy Reward Machines, the output symbols for $L^*$ are equated with rewards in $\mathbb{R}$.
\begin{definition}[Mealy Reward Machine]
\label{def:RT-Rens}
Given a set of states $S$, a set of actions $A$ and a labeling function $\lambda$, a \emph{Mealy Reward Machine} (MRM) is a tuple $\langle U, u_0, Z, \delta_u, \delta_r \rangle$, where
\begin{itemize}
\item $U$ is a finite set of MRM nodes,
\item $u_0 \in U$ is the start node,
\item $Z$ is a set of observations,
\item $\delta_u:U\times Z \mapsto U$ is the transition function, such that $\delta_u(u_i, \lambda(a,s))=u_j$ for $a\in A$ and $s\in S$,
\item $\delta_r:U\times Z \mapsto \mathbb{R}$ is the reward-output function, such that $\delta_r(u_i, \lambda(a,s)) = r'$ for $r'\in\mathbb{R}$, $a\in A$ and $s\in S$.
\end{itemize}
We may write $\delta_u^\mathit{MR}$ and $\delta_r^\mathit{MR}$ to emphasize that the functions are associated with MRM $\mathit{MR}$.
\end{definition}
A Markov Decision Process with a Mealy reward machine is thus an NMRDP.
In Figure~\ref{fig:MRM}, an edge from node $u_i$ to node $u_j$ labeled $\langle z\mid r \rangle$ denotes that $\delta_u(u_i,z)=u_j$ and $\delta_r(u_i,z)=r$. 

In the following definitions, let $s_i\in S$, $a_i\in A$, $z_i\in Z$, and $r_i\in\mathbb{R}$.
An \emph{interaction trace} of length $k$ in an MDP represents an agent's (recent) interactions with the environment. It has the form $s_0 a_0 r_1 s_1 a_1 r_2 \cdots s_k a_k r_{k+1}$ and is denoted $\sigma_\mathit{inter}$. That is, if an agent performs an action at time $t$ in a state at time $t$, then it reaches the next state at time $t+1$ where/when it receives a reward.
An \textit{observation trace} is extracted from an interaction trace (employing a labeling function) and is taken as input to an MRM. It has the form $z_1 z_2 \cdots z_k$ and is denoted $\sigma_z$.
A \textit{reward trace} is either extracted from an interaction trace or is given as output from an MRM. It has the form $r_1 r_2 \cdots r_{k+1}$ and is denoted $\sigma_r$.
A \textit{history} has the form $s_0 a_0 s_1 a_1 \cdots s_{k}$ and is denoted $\sigma_h$.
We extend $\delta_r$ to take histories by defining $\delta^*_{r}(u_i,\sigma_h)$ inductively as 
\[
\delta_r(u_i, \lambda(a_0,s_1))\cdot\delta^*_{r}(\delta_u(u_i, \lambda(a_0,s_1)),a_1 s_2 \cdots s_k).
\]
$\delta^*_{r}$ explains how an MRM assigns rewards to a history in an MDP.


A (deterministic) strategy to play in a nrMDP $M = \langle S, A, T, s_0 \rangle$ is a function $\pi:S^* \mapsto A$ that associates to all sequences of states $\sigma\in S^*$ of $M$ the action $\pi(\sigma)$ to play.

Given reward trace $\sigma_r = r_1 \cdots r_k$, \textit{discounted sum} and \textit{mean payoff} are defined as
\[
dSum^\gamma(\sigma_r) = \sum_{i=1}^k \gamma^i r_i  \mbox{ resp. } \mathcal{MP}(\sigma_r) = \frac{1}{k}\sum_{i=1}^k r_i.
\]
Let $\mathcal{UM}(\mathcal{R})$ be the discounted sum or mean payoff of an infinite reward trace generated by reward model $\mathcal{R}$.
Then the \textit{expected} discounted sum and mean payoff under strategy $\pi$ played in MDP $M$ from state $s$ is denoted as $\mathbb{E}^{M,\pi}_{s}(\mathcal{UM}(\mathcal{R}))$.




Being able to produce a traditional, immediate reward MDP from a non-Markovian rewards decision process is clearly beneficial: One can then apply all the methods developed for MDPs to the underlying NMRDP, whether to find optimal of approximate solutions. We produce an MDP from a non-reward MDP (nrMDP) and a Mealy reward machine by taking their product as defined next.

\begin{definition}
Given an nrMDP $M = \langle S, A, T, s_0 \rangle$, a labeling function $\lambda:S\mapsto Z\uplus\{null\}$ and an MRM $\mathit{MR}=\langle U, u_0, Z, \delta_u, \delta_r \rangle$, we define the synchronized product of $M$ and $\mathit{MR}$ under $\lambda$ as an (immediate reward) MDP $P = M \otimes_\lambda \mathit{MR} = \langle S^P, A^P, T^P, R^P, s_0^P \rangle$, where
$S^P = S\times U$,
$A^P = A$,
$T^P((s,u), a, (s',u'))=T(s, a, s')$ if $u' = \delta_u(u,\lambda(a,s'))$, else $T^P((s,u), a, (s',u'))=0$,
$R^P(a, (s,u)) = \delta_r(u,\lambda(a,s))$, and
$s_0^P = (s_0,u_0)$.
\end{definition}
%
The strategies in $M$ and $P$ are in bijection.
%
The following proposition states that the expected value of an nrMDP with an MRM is equal to the expected value of their product under the same strategy.
\begin{proposition}
\label{prp:expected-val-of-product-MDP}
Given an nrMDP $M = \langle S, A, T, s_0 \rangle$, a labeling function $\lambda:S\mapsto Z\uplus\{null\}$ and an MRM $\mathit{MR}=\langle U, u_0, Z, \delta_u, \delta_r \rangle$, for all strategies $\pi$ for $M$, we have that
\[
\mathbb{E}^{M,\pi}_{s_0}(\mathcal{UM}(\mathit{MR}_\lambda)) = \mathbb{E}^{P,\pi^P}_{(s_0,u_0)}(\mathcal{UM}(R^P)).
\]
\end{proposition}
Because memoryless strategies are sufficient to play optimally in order to maximize $\mathbb{E}^{P,\pi^P}_{(s_0,u_0)}(\mathcal{UM}(R^P))$ in immediate reward MDPs, and together with Proposition~\ref{prp:expected-val-of-product-MDP}, we conclude that we can play optimally in $M$ under $\mathit{MR}$ in a finite memory strategy (the memory required for $\mathit{MR}$; but memoryless if viewed as $P = M \otimes_\lambda \mathit{MR}$).

\section{Learning Mealy Reward Machines}
\label{sec:Learning-Mealy-Reward-Machines}

We make the important assumption that the environment and the agent's state can be reset. Recall the Treasure-Map world. After receiving the treasure map, the agent might not find a guide or equipment. In general, an agent might not finish a task or finish only one version of a task or subtask. Our reset assumption allows the agent to receive map again and explore various trajectories in order to learn the entire task with all its variations. Resetting also sets the underlying and hypothesized MRMs to their start nodes. Of course, the agent retains the hypothesized MRM learnt so far. Resetting a system is not always feasible, however, we are interested in domains where an agent can be placed in a position to continue learning or repeat its task.

\subsection{Learning and Exploitation}
\label{sec:Optimal-Learning-and-Exploitation}

\paragraph*{Problem Statement.}
Given an nrMDP $M = \langle S, A, T, s_0 \rangle$, a labeling function $\lambda:S\mapsto Z\uplus\{null\}$, an  \textit{unknown} MRM $\mathit{MR}=\langle U, u_0, Z, \delta_u, \delta_r \rangle$ which needs to be learned, and a threshold $\theta$, learn \textit{whenever possible}, a finite memory strategy $\pi^\theta$ such that $\mathbb{E}^{M,\pi}_{s_0}(\mathcal{UM}(\mathit{MR}_\lambda)) \geq \theta$.

There are several ways to determine $\theta$. One straightforward way is to deploy the agent in the environment in a pre-trial phase
and set $\theta$ to the highest return achieved. Another way is to set $\theta$ dynamically: let $\rho$ be the highest return observed so far; set $\theta = \rho + \beta\dot{\rho}$, where $\dot{\rho}$ is the a measure of the change in $\rho$ and $\beta\in [0,1]$ is a weighting factor related to the confidence that $\rho$ will keep changing by $\dot{\rho}$.

We observe that although an agent may find an optimal strategy for which $\mathbb{E}^{M,\pi}_{s_0}(\mathcal{UM}(\mathit{MR}_\lambda)) \geq \theta$, if the MDP is not strongly connected, the agent could end up in a region of the state space where it is impossible to gain rewards greater than $\theta$. For this reason, we shall make the agent reset if it has not yet received $\theta$ or more rewards after a fixed number of actions executed in an epoch.
A high-level description of the process for solving the problem follows.
For learning $\pi^\theta$, we consider an active learning scenario in which the agent can
\begin{enumerate}
\item play a finite number of episodes to answer membership queries until the observation tree (OT) is complete by extracting reward traces from the appropriate interaction traces;
\item construct new hypothesized MRM $\mathcal{H}$ from OT and compute the optimal strategy $\pi^*$ for $M \otimes_\lambda \mathcal{H}$, as soon as OT becomes complete;
\item start performing actions to complete its task, using $\pi^*$ until a counter-example to $\mathcal{H}$ is discovered (experienced) if the expected value of $\pi^*$ is greater than $\theta$, in which case, stops exploitation and go to step 1 (the learning phase); else use conformance testing techniques to refute $\mathcal{H}$ if the expected value of $\pi^*$ is less than $\theta$, and go to step 1;
\item restart its task if a given number of actions have been played and its total rewards (w.r.t.\ $\mathcal{UM}$) is still less than $\theta$.
\end{enumerate}
 
Algorithm~\ref{alg:optimal-active-learning} is the general, high-level algorithm for an agent actively learning an MRM in an MDP, and acting optimally with respect to the currently hypothesized MRM $\mathcal{H}$.
In the algorithm, ``alive'' implies that there is no specific stopping criterion. In practice, the user of the system/agent will specify when the system/agent should stop collecting rewards.
Procedures getMQ$(OT)$, resolveMQ$(OT, \sigma_z, \sigma_r)$, buildRewardMachine$(OT)$ and addCounterExample$(OT,$ $\sigma_z, \sigma_r)$ are part of the Mealy machine inference algorithm which we assume to be predefined.
%

The getExperience$(\sigma_z, s_0)$ procedure hides much important detail. In this procedure, the agent performs actions in order to answer the membership query starting from state $s_0$. There are potentially many ways for the agent to behave, from performing random exploration to seeking an optimal strategy to encounter the sequence of observations represented by $\sigma_z$. We leave the investigation of guarantees for strategies for reaching a $\sigma_z$ in an MDP from a give state for the future. In the next section, we propose one reasonable, approximate strategy.

\begin{algorithm}[h!]
\begin{normalsize}
\caption{Optimal Active Learning
\label{alg:optimal-active-learning}}
\begin{algorithmic}
\STATE Initialize observation table $OT$
\WHILE{alive}
	\IF{$OT$ \textnormal{is not complete}}
		\STATE $\sigma_z \gets$ getMQ$(OT)$\;
		\STATE $\sigma_r \gets$ getExperience$(\sigma_z, s_0)$\;
		\STATE resolveMQ$(OT, \sigma_z, \sigma_r)$\;
	\ELSE
		\STATE $\mathcal{H} \gets$ buildRewardMachine$(OT)$\;
		\STATE $\pi^*\gets \argmax_{\pi}\mathbb{E}^{M,\pi}_{s_\mathit{cur}}(\mathcal{UM}(\mathcal{H}))$\;
		\IF{$\max_{\pi}\mathbb{E}^{M,\pi}_{s_\mathit{cur}}(\mathcal{UM}(\mathcal{H})) \geq \theta$}
			\STATE $\pi^\theta\gets\pi^*$\;
			\STATE actionsExecuted $\gets 0$\;
			\STATE $s_\mathit{cur} \gets s_0$\;
			\STATE $u_\mathit{cur} \gets u_0^\mathcal{H}$\;
			\REPEAT
				\STATE $a\gets \pi^\theta((s_\mathit{cur}, u_\mathit{cur}))$\;
				\STATE $s_\mathit{nxt}\thicksim T(s_\mathit{cur}, a, \cdot)$\;
				\STATE $u_\mathit{nxt}\gets\delta_u^\mathcal{H}(u_\mathit{cur},\lambda(a, s_\mathit{nxt}))$\;
				\STATE $r\gets \delta_u^\mathcal{H}(u_\mathit{cur},\lambda(a, s_\mathit{nxt}))$\;
				\STATE Increment actionsExecuted by 1\;
				\STATE Update $\sigma_\mathit{inter}$ with $(a, s_\mathit{nxt}, r)$\;
				\STATE $\sigma_r\gets$ extractRewTrace$(\sigma_\mathit{inter})$\;
				\IF{$\mathcal{UM}((\sigma_r)) < \theta$ \textnormal{and actsExtd} $\geq$ \textnormal{ActsToExt}}
					\STATE $s_\mathit{cur} \gets s_0$\;
					\STATE $u_\mathit{cur} \gets u_0^\mathcal{H}$\;
				\ELSE
					\STATE $s_\mathit{cur} \gets s_\mathit{nxt}$\;
					\STATE $u_\mathit{cur} \gets u_\mathit{nxt}$\;
				\ENDIF
			\UNTIL{$\sigma_\mathit{inter}$ \textnormal{is a counter example to} $\mathcal{H}$}
				\STATE $\sigma_z\gets$ extractObsTrace$(\sigma_\mathit{inter})$\;
           		\STATE $\sigma_r\gets$ extractRewTrace$(\sigma_\mathit{inter})$\;
            	\STATE addCounterExample$(OT, \sigma_z, \sigma_r)$\;
		\ELSE
			\STATE conformanceTesting$(\mathcal{H})$\;
		\ENDIF
	\ENDIF
\ENDWHILE
\end{algorithmic}
\end{normalsize}
\end{algorithm}

Procedure conformanceTesting$(\mathcal{H})$ uses conformance testing techniques to obtain an interaction trace $\sigma_\mathit{inter}$ that shows a discrepancy between the underlying $\mathit{MR}_\lambda$ and $\mathcal{H}$, and use $\sigma_\mathit{inter}$ to resolve $OT$\;

%

\paragraph{On guarantees offered by our learning setting}
The main guarantees offered by the building blocks of our algorithm are as follows.

First, let us comment on the guarantees offered by the $L^*$ algorithm in our particular context. As we already mentioned, the realization of membership queries through the execution of possibly multiple episodes in the MDP is already a non-trivial task that cannot be realized with absolute certainty. Indeed, if $L^*$ asks, as a membership query, for the reward obtained after the sequence of observations $\sigma_z=z_1 z_2 \dots z_n$, we need to consider two cases. $(a)$ This sequence may not exist in the MDP $M$, in that case, we can provide $L^*$ with an arbitrary reward because this reward will never be used when considering how to play in $M \otimes_{\lambda} \mathit{MR}$. $(b)$. If such a sequence does exists, let $\sigma_a=a_1 a_2 \dots a_n$ be a sequence of actions that maximizes the probability $\alpha$, $0 < \alpha \leq 1$, of observing $\sigma_z$ from the initial state of $M$. If $\alpha <1$, repeating this sequence does guarantee that we will eventually observe $\sigma_z$ and thus its associated reward, but only with probability one and not absolute certainty.  Also, while in principle, conformance testing is complete when a bound on the number of states is known for $\mathit{MR}$, in our setting the completeness can only be ensured with probability one and not absolute certainty as in the classical setting.

Second, the computation of optimal strategies for $M \otimes_{\lambda} \mathcal{H}$, once an hypothesis $\mathcal{H}$ has been formulated by the learning algorithm, can be solved exactly, both for the discounted sum and mean reward functions (using model-checking algorithms, \cite{bk08b}, e.g.). As a consequence, if the learning part of our algorithm returns an hypothesis that is correct for every sequences of observations $\sigma_z$ that can be executed in $M$, then our algorithm computes an optimal strategy with certainty.

\subsection{Approximate Learning and Exploitation}
\label{sec:Approximate-Learning-and-Exploitation}

In this section, we propose an active learning process for an NMRDP agent with a Mealy reward machine, where the strategies played are likely suboptimal. This approach is more suited to domains where optimality (e.g. a guarantee that $\max_{\pi}\mathbb{E}^{M,\pi}_{s_\mathit{cur}}(\mathcal{UM}(\mathcal{H})) \geq \theta$) is not important, but where reactiveness is more important. It is also the algorithm (Alg.~\ref{alg:approximate-active-learning}) we use for evaluation in Section~\ref{sec:Experimental-Evaluation}.

\paragraph{Action Planning for Answering Membership Queries}

Something we have not yet discussed in detail is what strategy an agent follows to answer a membership query, that is, how procedure getExperience$(\sigma_z, s_0)$ in the algorithms is implemented. We now describe how the procedure is implemented for Algorithm~\ref{alg:approximate-active-learning}.
Whenever the agent receives a membership query, it starts planning and executing actions to reach a state where the first observation $z_1\in \sigma_z$ is made in $s_1$. The reward $r_1$ received for entering $s_1$ is recorder. Immediately, the agent plans and executes actions to reach $s_2$ where the second observation $z_2\in \sigma_z$ is made. The reward $r_2$ received for entering $s_2$ is recorder. This process continues until $\sigma_z$ has been observed and $r_1 r_2 \ldots r_k$ has been received (and recorded). $\sigma_r=r_1 r_2 \ldots r_k$ is then given to $L^*$ (via resolveMQ) as answer to the current membership query.

$Plan(\mathcal{H}, s_\mathit{cur})$ in Algorithm~\ref{alg:approximate-active-learning} is instantiated as Monte Carlo Tree Search \cite{bpwlctpsc12} planning. The planner employs reward function
\[
R(a, s, z) = \left\lbrace
\begin{array}{rl}
-x & \mbox{if } \lambda(a,s) = null\\
-y & \mbox{else if } \lambda(a,s) \neq z\\
y & \mbox{otherwise},
\end{array}
\right.
\]
where $z \in \sigma_z$ is the observation current being pursued by the agent and $y > x$.

\paragraph{Action Planning for Exploitation}

Once an hypothesis reward model is available, the agent can stop learning and start exploiting the model.

Assume that all the agent's tasks can be accomplished within $k$ non-null observations.
Let $Z^k$ be all non-null observation sequences of length $k$ and $R^k$ be all reward sequences of length $k$.
We seek the sequence $\sigma^*_z\in Z^k$ of observations that will maximize the sum of rewards it induces in $\mathcal{H}$. 
Let $H^k$ be all histories with $k$ action-state pairs.
Let $\sigma_h = s_0 a_0 s_1 a_1 \cdots s_{k} \in H^k$.
We mean by $\sigma_z(\sigma_h)$ the observation sequence induced by history $\sigma_h$. That is, $\sigma_z(\sigma_h) = \lambda(a_0,s_1)\cdots \lambda(a_{k-1} s_k)$.
Then, we seek $\sigma^*_z(\sigma_h^*)$ where
$
\sigma_h^* = \argmax_{\sigma_h\in H^k}\sum_{r\in \delta^*_{r}(u_0,\sigma_h)}r.
$
This is the sequence returned by getGoodObsSeq$(\mathcal{H},$ $k)$ in Algorithm~\ref{alg:approximate-active-learning}.

With $\sigma^*_z$ in hand, the agent simply plans (using $Plan$) and acts in order to experience $\sigma^*_z$.
The sequence $\sigma^*_z$ is not guaranteed to be optimal because it does not consider state transition probabilities. But it is sufficient for our implementation for our proof of concept evaluation. If, while exploiting, the agent experiences a trace that contradicts the hypothesized reward model, it should go back to the learning phase.

\begin{algorithm}[h!]
\begin{normalsize}
\caption{Approximate Active Learning
\label{alg:approximate-active-learning}}
\begin{algorithmic}
\STATE Initialize observation table $OT$\;
\WHILE{alive}
	\IF{$OT$ \textnormal{is not complete}}
		\STATE $\sigma_z \gets$ getMQ$(OT)$\;
		\STATE $\sigma_r \gets$ getExperience$(\sigma_z, s_0)$\;
		\STATE resolveMQ$(OT, \sigma_z, \sigma_r)$\;
	\ELSE
		\STATE $\mathcal{H} \gets$ buildRewardMachine$(OT)$\;
		\STATE $\sigma_z^k\gets$ getGoodObsSeq$(\mathcal{H}, k)$\;
		\STATE actsExtd $\gets 0$\;
		\STATE $s_\mathit{cur} \gets s_0$\;
		\STATE $u_\mathit{cur} \gets u_0^\mathcal{H}$\;
		\REPEAT
			\STATE $s_\mathit{cur}\gets random(S)$\;
			\FOR{$z\in\sigma_z^k$}
				\REPEAT
					\STATE $a\gets Plan(\mathcal{H}, s_\mathit{cur})$\;
					\STATE $s_\mathit{nxt}\thicksim T(s_\mathit{cur}, a, \cdot)$\;
					\STATE $r\gets \delta_u^\mathcal{H}(u_\mathit{cur},\lambda(a, s_\mathit{nxt}))$\;
					\STATE Increment actsExtd by 1\;
					\STATE Update $\sigma_\mathit{inter}$ with $(a, s_\mathit{nxt}, r)$\;
					\STATE $s_\mathit{cur} \gets s_\mathit{nxt}$\;
					\STATE $u_\mathit{cur} \gets u_\mathit{nxt}$\;
				\UNTIL{$\lambda(a,s_\mathit{cur}) = z$ \textnormal{or actsExtd} $\geq$ \textnormal{actsToExt}}
			\ENDFOR
		\UNTIL{$\sigma_\mathit{inter}$ \textnormal{is a counter example to} $\mathcal{H}$ \textnormal{and actsExtd} $<$ \textnormal{actsToExt}}
		\IF{\textnormal{actsExtd} $<$ \textnormal{actsToExt}}
            \STATE $\sigma_z\gets$ extractObsTrace$(\sigma_\mathit{inter})$\;
            \STATE $\sigma_r\gets$ extractRewTrace$(\sigma_\mathit{})$\;
            \STATE addCounterExample$(OT, \sigma_z, \sigma_r)$\;
        \ENDIF
	\ENDIF
\ENDWHILE
\end{algorithmic}
\end{normalsize}
\end{algorithm}

\section{Experimental Evaluation}
\label{sec:Experimental-Evaluation}

Approximate active learning (Alg.~\ref{alg:approximate-active-learning}) was implemented.

\subsection{Experiments with the Treasure-Map world}

For learning and planning in the Treasure-Map world, the agent maintains an interaction-trace of length 6 (i.e., $k=6$). When Monte Carlo Tree Search is used, trajectories are simulated for 30 actions, and there are 100 trajectories per action being planned for. Every trial allows an agent 2000 exploitation actions; actions required for answering membership queries are free.

The agent can move north, east, west and south, one cell per action. To add stochasticity to the transition function, the agent is made to get stuck a percentage of the time when executing an action. We abbreviate the precision factor of actions as APF; for instance, if APF = $0.9$, then the agent gets stuck $10\%$ of the time. The default reward is set to $c=-1$ (the agent gets $-1$ when performing an action and seeing $null$). This is also the output/reward for all self-looping transitions in the hypothesis MRM.

We measure the total rewards gained per trial (Return), number of attempts to membership queries ($\#$MQAs), number of counter-examples encountered ($\#$CEs), total amount of time used for learning (LT; in seconds), number of exploration epochs/resets per trial ($\#$Epochs) and total amount of time used for exploitation (ET).
Tables~\ref{tbl:learning} and \ref{tbl:exploiting} show the average results over ten trials of each of three choices for the action precision factor.

The results in Table~\ref{tbl:learning} focus on learning. The exploitation strategy is to perform random actions. The return is thus not important for this experiment but is used as baseline for the experiments focusing on exploitation. In every trial, the MRM in Figure~\ref{fig:MRM} is correctly learnt.

The results in Table~\ref{tbl:exploiting} focus on exploitation. The exploitation strategy is to select actions with Monte Carlo Tree Search (MCTS) planning as described in Section~\ref{sec:Approximate-Learning-and-Exploitation}. The return is thus important.
Although the return varies with the action precision factor, the MRM in Figure~\ref{fig:MRM} is correctly learnt in every trial. This is because the learning process depends mostly on answering membership queries; these queries are always answered, even if it takes longer to answer a query due to low APF.

\begin{table}
\caption{Results focusing on the learning phase.
\label{tbl:learning}
}
\centering
\begin{tabular}{|c|c|c|c|c|}
\hline
APF	& $\#$MQAs				& LT (s) 					&	$\#$CEs				& Return \\
\hline
0.75	& $244\pm 68$		& $493 \pm 23$	& $0.4 \pm 0.5$	& $-1844\pm 144$ \\
\hline
0.85 & $341\pm 103$	& $496 \pm 9$	& $0.9 \pm 0.9$ & $-1830 \pm 105$ \\
\hline
0.95 & $244\pm 68$		& $443 \pm 30$	& $0.7 \pm 0.7$ & $-1794 \pm 103$ \\
\hline
\end{tabular}
\end{table}
\begin{table}
\caption{Results focusing on the exploitation phase.
\label{tbl:exploiting}
}
\centering
\begin{tabular}{|c|c|c|c|c|}
\hline
APF 	& Return					& ET	 (s)					& $\#$Epochs		& $\#$CEs \\
\hline
0.75 & $495 \pm 309$ & $131 \pm 12$	& $12 \pm 0.8$	& $7.9 \pm 1.6$ \\
\hline
0.85 & $1087 \pm 538$ & $144 \pm 19$ & $15 \pm 1$		& $9.6 \pm 2.4$ \\
\hline
0.95 & $1459 \pm 376$ & $174 \pm 59$ & $17 \pm 1$ & $10.6 \pm 1.3$ \\
\hline
\end{tabular}
\end{table}

\subsection{Learning in the Cookie Domain}

\begin{figure}[t]
\centering
\includegraphics[scale=0.25]{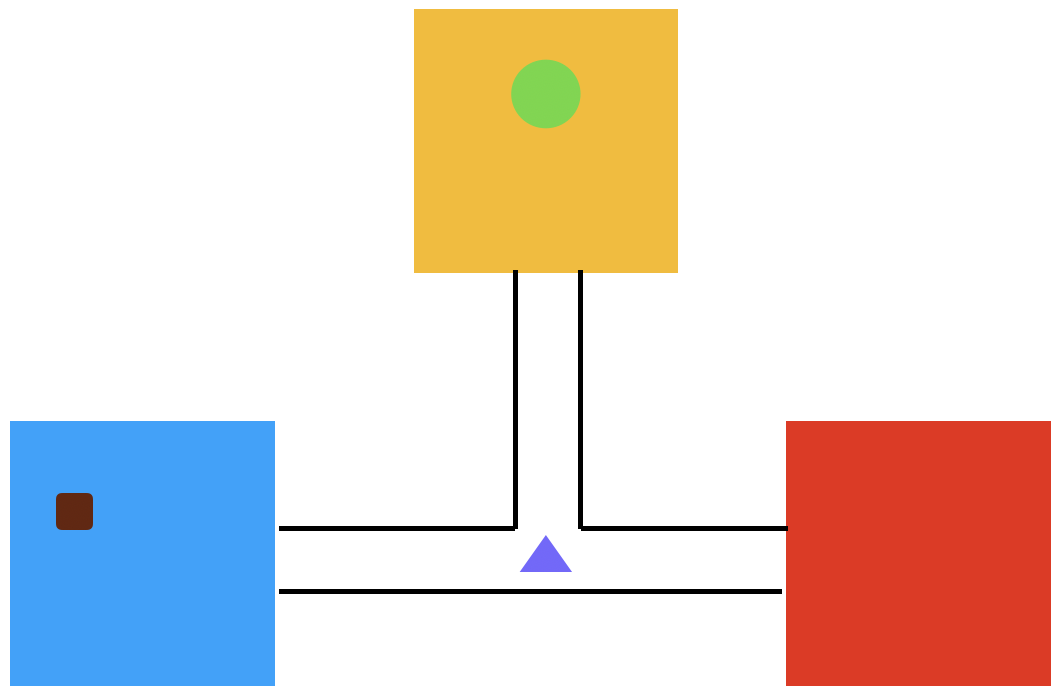}
\caption{The Cookie Domain. The button in yellow room causes a cookie to appear randomly either in the blue or red room. Agent: purple triangle.
\label{fig:cookie-domain}}
\end{figure}

\begin{figure}[t]
\centering
\includegraphics[scale=0.3]{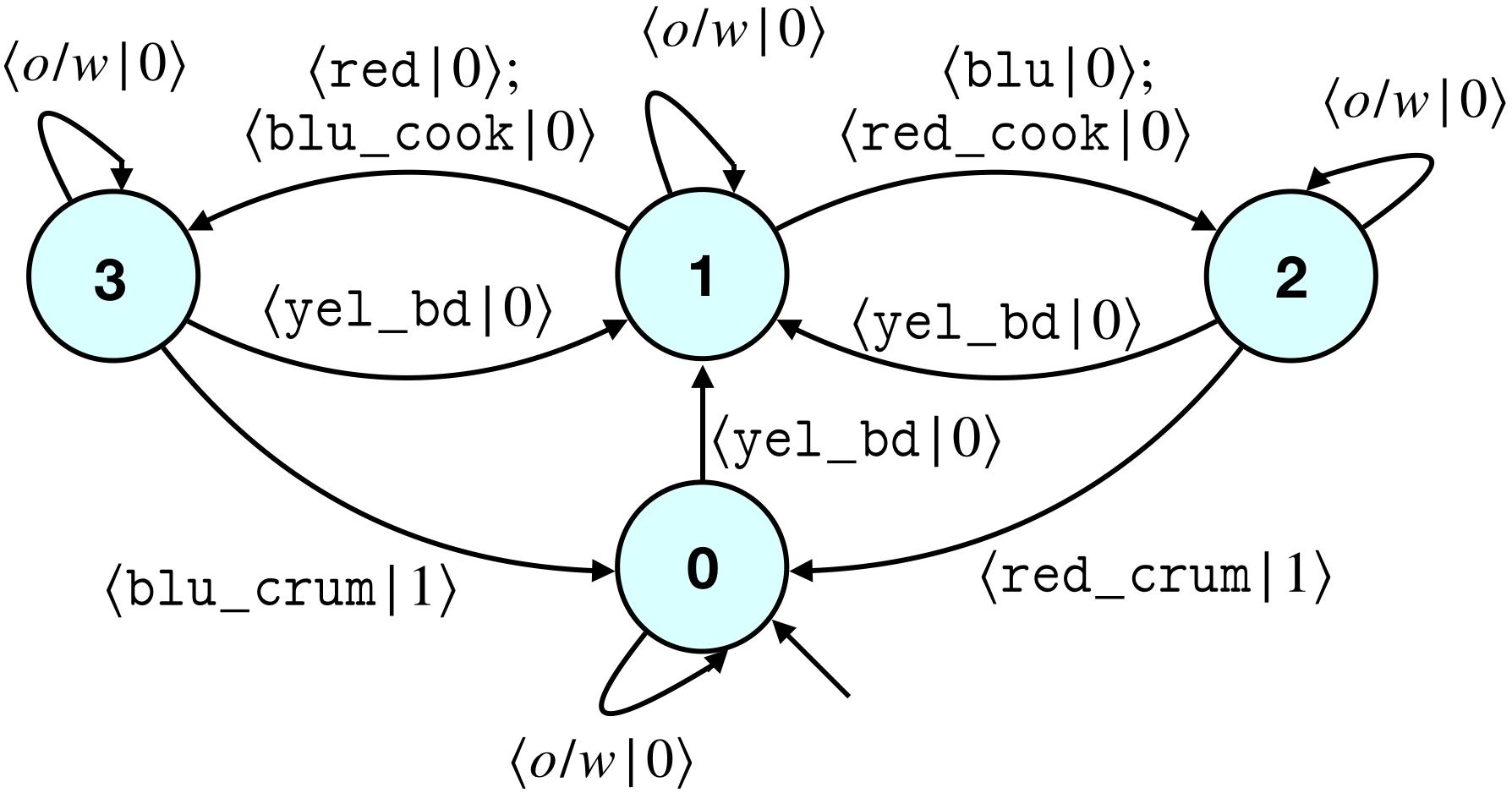}
\caption{The Cookie Domain (Mealy) reward machine.
\label{fig:cook-dom-RM}}
\end{figure}

The Cookie Domain of \cite{twkvcm19} is depicted in Figure~\ref{fig:cookie-domain}. The agent can press a button in the yellow room for a cookie to appear in either the blue or red room (which room is chosen randomly). If the button is pressed, a fresh cookie appears at random. The agent gets 1 reward for eating the cookie. This is a partially observable MDP (POMDP) because the agent cannot see where the cookie is (if there is one) until it is in the same room as the cookie.

In the original problem \cite{twkvcm19}, the agent can move in the four cardinal directions and push the button. When the agent enters a room where a cookie is, it automatically eats it. We added an explicit \textit{eat} action, which leaves crumbs in the room if the agent is in the same room as a cookie. All actions and observations are deterministic. They have a set of propositions $\{\mathtt{blue},$ $\mathtt{red},$ $\mathtt{yellow},$ $\mathtt{cookie},$ $\mathtt{crumbs},$ $\mathtt{button\_pressed}\}$ which we interpret as observations. However, our particular observations are $\{\mathtt{blu},$ $\mathtt{blu\_cook},$ $\mathtt{blu\_crum},$ $\mathtt{red},$ $\mathtt{red\_cook},$ $\mathtt{red\_crum},$ $\mathtt{yel},$ $\mathtt{yel\_bd}\}$, meaning the agent is in the empty blue room, the blue room with a cookie, the blue room with crumbs, the empty red room, the red room with a cookie, the red room with crumbs, the yellow room with the button not pressed down and the yellow room with the button depressed. 

We defined our labeling function to simply report what the case is in state $s$ (room color, cookie, crumbs), and if the agent is in the yellow room and $a$ is $\mathtt{push\_button}$, then $\mathtt{yel\_bd}$ is returned, else $\mathtt{yel}$ is returned. The function returns $null$ if the agent is in the passage between the rooms.

The authors provide the reward machine depicted in Figure~\ref{fig:cook-dom-RM} as a ``perfect RM''. We provided it as the underlying MRM for our experiment. For this experiment, we implemented getExperience simply as random exploration. We performed ten trials. Our agent learns the 'perfect (M)RM' perfectly every time, requiring approximately one second per trial.

\section{Conclusion}

We proposed two frameworks for learning and exploiting non-Markovian reward models. The reward models are represented as Mealy machines. Angluin's $L^*$ algorithm was employed to learn \textit{Mealy Reward Machines} within a Markov Decision Process setting. The one framework was justified theoretically, while the other framework was based on an actual implementation that was used for evaluation and as a proof-of-concept.
We found that the latter framework always learns the toy examples correctly, that is, after answering a finite number of membership queries as posed by the $L^*$ algorithm, within a reasonable time.

When in a (fully observable) MDP, an MRM might be used for avoiding detrimental situation that cannot be avoided in an MDP with a traditional immediate reward function. Observe that, in a sense, an MRM maps a sequence of actions and states visited to a particular reward; this cannot be done with a traditional immediate reward function. Consider, for instance, the state where a battery pack of an electric vehicle (EV) is finished being assembled. Imagine that there is a crucial process that must be avoided, else the battery pack could explode during use in the EV. Sequences of assembly including this dangerous process could be assigned a very low reward. If it can be shown that every sequence involving the bad process results in a reward less than some threshold, then any sequence resulting in a reward greater than the threshold is guaranteed to be \textit{safe}.

In future, we would like to implement the framework in a more principled way, using the optimal approach as a starting point. 
We expect that larger underlying reward machines will require intelligent exploration strategies. It will be interesting to investigate the exploration-exploitation trade-off in the setting of non-Markovian rewards. We would also like to compare our work more closely to that of \cite{twkvcm19}, but at the time of this research, their implementation was not available.

\bibliography{references}
\bibliographystyle{ecai}
\end{document}